\documentclass[letterpaper, 10 pt, conference]{ieeeconf}  

\IEEEoverridecommandlockouts                              

\usepackage{graphics} 
\usepackage{epsfig} 
\usepackage{mathptmx} 
\usepackage{times} 
\usepackage{amsmath} 
\usepackage{amssymb}  
\usepackage{tikz} 
\usetikzlibrary{positioning}
\usepackage{subfigure}

\usepackage[noadjust]{cite}

\def\eg{\emph{e.g.}} 

\def\ie{\emph{i.e.}}

\makeatother

\title{\LARGE \bf
CNN-Augmented Visual-Inertial SLAM with Planar Constraints
}

\author{Pan Ji, Yuan Tian, Qingan Yan, Yuxin Ma, and Yi Xu 

\thanks{The authors are with OPPO US Research Center, Palo Alto, CA 94303, USA.
        {\tt\small firstname.lastname@oppo.com}}%
}

\begin{document}

\maketitle
\thispagestyle{empty}
\pagestyle{empty}

\begin{abstract}
We present a robust visual-inertial SLAM system that combines the benefits of Convolutional Neural Networks (CNNs) and planar constraints. Our system leverages a CNN to predict the depth map and the corresponding uncertainty map for each image. The CNN depth effectively bootstraps the back-end optimization of SLAM and meanwhile the CNN uncertainty adaptively weighs the contribution of each feature point to the back-end optimization. Given the gravity direction from the inertial sensor, we further present a fast plane detection method that detects horizontal planes via one-point RANSAC and vertical planes via two-point RANSAC. Those stably detected planes are in turn used to regularize the back-end optimization of SLAM. We evaluate our system on a public dataset, \ie, EuRoC, and demonstrate improved results over a state-of-the-art SLAM system, \ie, ORB-SLAM3.
\end{abstract}

\section{INTRODUCTION}
Simultaneous Localization and Mapping (SLAM) aims to reconstruct the 3D map of a scene and localize the position of an agent with respect to the map at the same time. It has been extensively researched over the years~\cite{cadena2016past} and has found various applications, such as autonomous navigation, Virtual Reality (VR), and Augmented Reality (AR).

The sensor types that SLAM can work with include monocular/stereo cameras, depth sensor, and the inertial measurement unit (IMU). The monocular mode is the minimal setup for visual SLAM but does not produce scale information. Stereo and RGB-D modes can provide metric scale recovery for map points and camera poses; however, extra sensor requirements (\eg, stereo cameras and depth sensors) hinder their wide applicability. Nowadays, with inertial sensors becoming cheaper and easily accessible via devices such as mobile phones, visual-inertial SLAM has become a popular option for self-localization, \eg, in mobile-based AR applications~\cite{qin2018vins}, as inertial sensors can help SLAM achieve metric-scale reconstruction. Nonetheless, visual-inertial SLAM still suffers from scale drift because the inertial sensor inevitably accumulates errors over time~\cite{qin2018vins}. 

On the other hand, depth prediction from a single image using CNNs has also achieved great success~\cite{fu2018deep,godard2019digging}, demonstrating complimentary features to traditional geometric methods. Hence, a few works~\cite{tiwari2020pseudo,yang2020d3vo,loo2019cnn,tateno2017cnn,yang2018deep} try to combine the best of both worlds, \ie, deep-learning-based depth prediction and geometry-based SLAM. However, they do not incorporate the inertial sensors that are currently easy to access and the structural constraints that are commonly presented in 
real-world scenes.


In this paper, we target an accurate and robust visual-inertial SLAM system that uses a monocular camera and an IMU sensor to capture input data. To this end, we propose to augment the visual-inertial SLAM with CNNs that predict the depth map and the uncertainty map for each image, forming a pseudo RGB-D inertial SLAM system. As such our SLAM system combines the best of geometry and deep learning, albeit still using a minimal sensor set-up. Furthermore, since gravity direction can be estimated from inertial measurements, we can fit the dominant planes much faster with fewer points required to sample. 
We further enhance the robustness of plane detection leveraging the depth and uncertainty estimated by the CNN, and in turn use the detected planes to regularize the back-end optimization of SLAM.

In summary, our contributions include:
\begin{itemize}
    \item[(a)]	a robust visual-Inertial SLAM system augmented with CNN depth and uncertainty;
    \item[(b)]	a fast and robust plane fitting method using gravity direction from inertial measurements, map points from SLAM, and depth and uncertainty from the CNN;
    \item[(c)]	a simple yet effective plane regularization energy term for SLAM back-end optimization.
\end{itemize}

\section{RELATED WORK}
\subsection{Geometric SLAM}
SLAM has been a long standing problem in the community of robotics and computer vision. Existing geometric SLAM systems can be roughly divided into two categories: feature-based SLAM~\cite{song2013parallel,mur2015orb} and direct SLAM~\cite{engel2014lsd,engel2017direct}. Feature-based SLAM systems first detect feature points and then compute the geometry using feature correspondences. The performance of those systems degrades significantly when feature detection and tracking work poorly, \eg, in scenes with little texture. In contrast, direct SLAM systems rely on the photometric error, \ie, the difference of pixel intensities, to optimize the geometry, and thus bypass the need of extracting feature points and computing feature descriptions. However, direct methods assume brightness constancy 
and suffer a lot in environments that have rapid lighting changes. 

The robustness of SLAM can be improved using multi-modality sensors, such as the RGB-D camera and the IMU sensor. Along this line, the ORB-SLAM family of methods~\cite{mur2015orb,mur2017orb,campos2020orb} gradually support more sensors. Starting from the monocular ORB-SLAM~\cite{mur2015orb}, ORB-SLAM2~\cite{mur2017orb} adds both stereo and RGB-D modes, and ORB-SLAM3~\cite{campos2020orb} further incorporates the inertial sensor, demonstrating the state-of-the-art performance. Due to its robustness and versatility, we build our system upon ORB-SLAM3.

There have also been other efforts~\cite{lee2011mav,arndt2020points,hsiao2018dense,kaess2015simultaneous,yang2016pop,hosseinzadeh2018structure} that exploit structural information, \eg, planar constraints, to improve the performance of geometric SLAM. A few of them~\cite{lee2011mav,arndt2020points,zhou2020efficient} detect and fit 3D planes within feature-based monocular SLAM, which have difficulties in finding large textureless planes. Others~\cite{hsiao2018dense,kaess2015simultaneous,yang2016pop,hosseinzadeh2018structure} enforce planar constraints for RGB-D SLAM; thus achieving more reliable results. However, none of them utilize the inertial sensor to simplify the plane fitting problem. In our work, we combine the inertial measurements, sparse map points, dense CNN depth and uncertainty, to build a fast and robust method for plane detection and fitting.

\subsection{CNN-Augmented SLAM}
Depth prediction using deep CNNs has recently achieved great success~\cite{fu2018deep,godard2019digging,tiwari2020pseudo,zou2020learning,ji2021monoindoor,liu2022planemvs,ji2022georefine}. Hence, a few works~\cite{tiwari2020pseudo,yang2020d3vo,loo2019cnn,tateno2017cnn,yang2018deep} try to combine deep learned depth/pose and geometric SLAM. In particular, \cite{tateno2017cnn}, \cite{loo2019cnn} and \cite{yang2020d3vo} leverage deep depth, pose and/or uncertainty to bootstrap a semi-dense visual odometry system. \cite{tiwari2020pseudo} uses the CNN depth as the input to an RGB-D SLAM system and builds a self-improving loop for depth prediction and monocular SLAM. However, none of them incorporates the inertial sensors, which are widely available on mobile devices, and the planar constraints, which commonly exist in indoor environments. In contrast, our system is incarnated with deep neural networks, visual-inertial SLAM, plane detection, and planar constraints.

\section{METHOD}

In this section, we present our CNN-augmented visual-inertial SLAM and show how we can detect and fit 3D planes fast and robustly under our framework. 

\subsection{Overall Pipeline}
\begin{figure}[!ht] 
  \centering
  \includegraphics[width=0.95\linewidth]{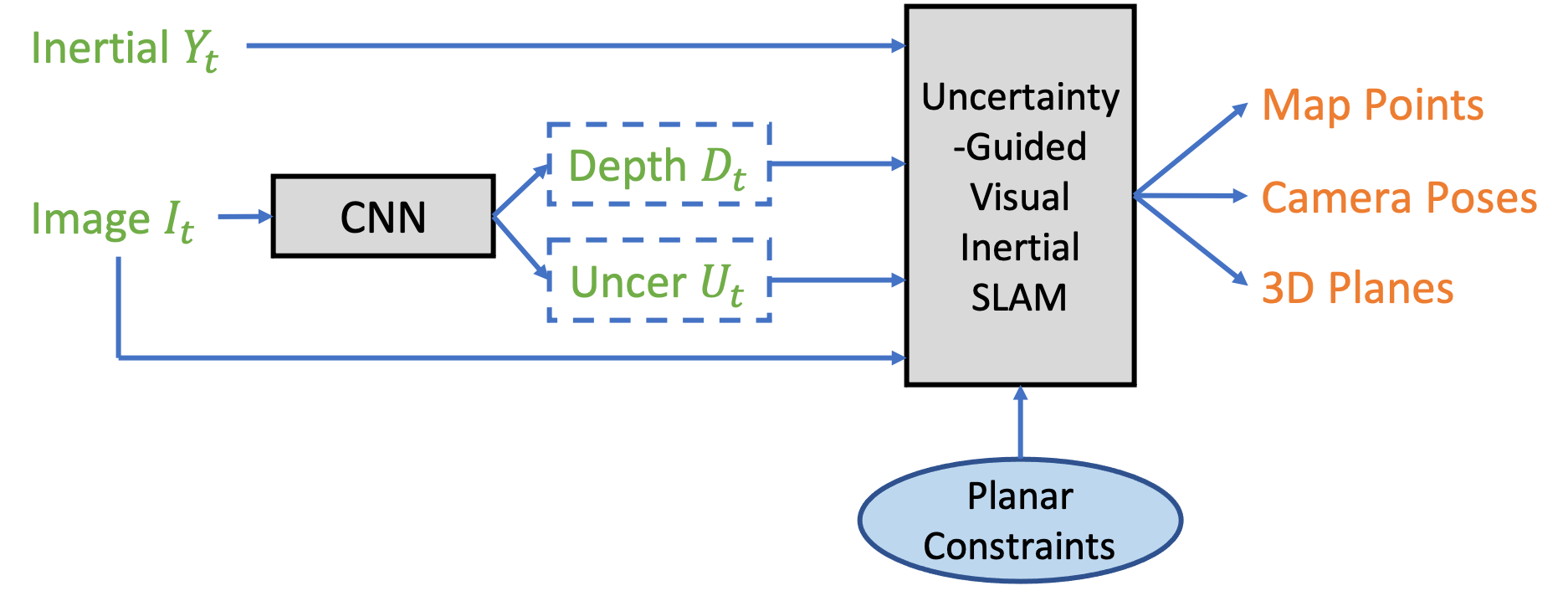}
  \caption{Overall pipeline of the proposed CNN-augmented visual-inertial SLAM with planar constraints.} 
  \label{fig:pipeline}
\end{figure}

Our visual-inertial SLAM combines conventional geometric SLAM and modern CNN-based depth/uncertainty predictions, and further enforces planar constraints to improve performance. The overall pipeline is depicted in Figure~\ref{fig:pipeline}. We first train a CNN to predict the depth map $D_t$ and its uncertainty map $U_t$ from a single image $I_t$. The CNN can be trained in a supervised~\cite{kendall2017uncertainties} or self-supervised manner~\cite{yang2020d3vo}. The image $I_t$, its predicted depth map $D_t$, uncertainty map $U_t$, and inertial measurements $Y_t$ are then used as the input to an uncertainty-guided visual-inertial SLAM system. Planes are detected in a separate thread of the front-end. Planar constraints are enforced in the back-end through the factor graph optimization after a set of reliable planes are detected. The outputs of the system include map points and camera poses in the metric scale and steadily detected 3D planes.

\subsection{CNN Depth and Uncertainty}
\begin{figure}[!ht] 
  \centering
  \includegraphics[width=0.8\linewidth]{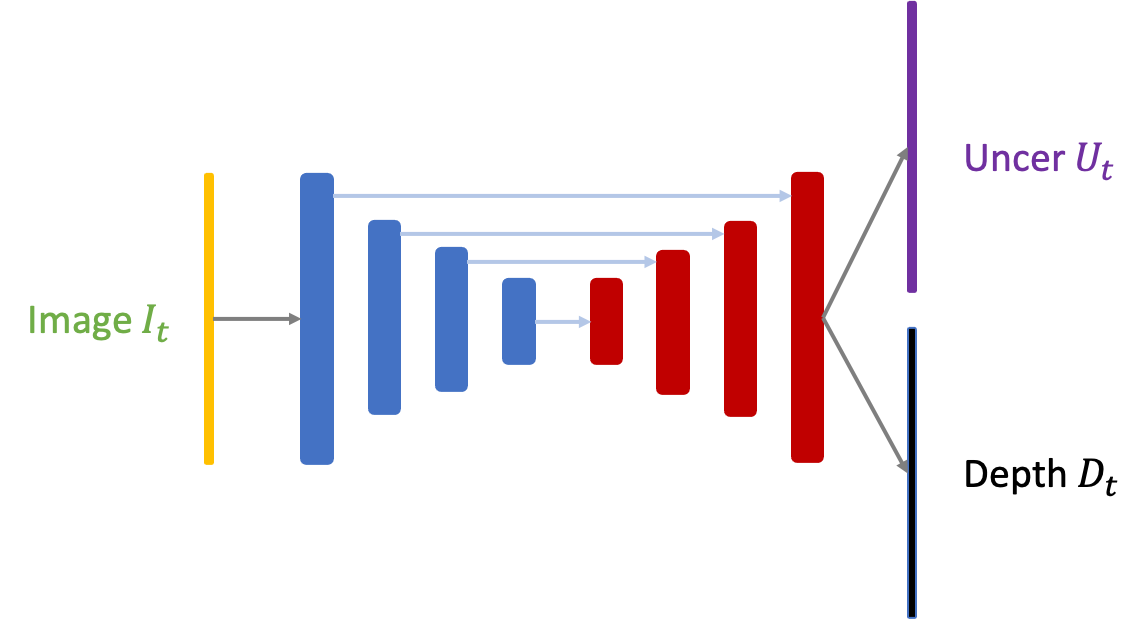}
  \caption{Network structure for CNN depth and uncertainty prediction. The network follows an auto-encoder structure with skip connections, and takes one image as input to predict its depth map and uncertainty map.} 
  \label{fig:cnn-depth-uncer}
\end{figure}

Deep learning is complementary to geometry in many scenarios. For instance, geometric SLAM tends to fail totally in textureless regions, whereas CNN-based methods can still predict reasonable depth values in such regions leveraging rich priors in the training data. So we leverage a CNN that predicts the depth and the uncertainty to augment the geometric SLAM. Following~\cite{yang2020d3vo}, we train the network in a self-supervised manner. However, note that our system is agnostic to the training strategy of the deep neural network and a supervised method shall also work. The network structure for depth and uncertainty prediction from an image is outlined in Figure~\ref{fig:cnn-depth-uncer}. The CNN takes an auto-encoder structure with skip-connections between the encoder and the decoder. We refer the reader to~\cite{yang2020d3vo} for the detailed network structure.  

CNN depth is not equally accurate in different regions of an image~\cite{kendall2017uncertainties}. We thus design the network to simultaneously predict the confidence or uncertainty score for the depth prediction. Specifically, we follow ~\cite{kendall2017uncertainties,yang2020d3vo} and define the uncertainty weighted self-supervised loss as follows,
\begin{equation}
L_{self}=\sum \frac{r_{pho}}{u}+\log{u},
\end{equation}
where $r_{pho}$ is the photometric loss as defined in~\cite{godard2019digging}, and $u$ is the uncertainty value predicted for each pixel. The total loss function is the summation of the uncertainty weighted self-supervised loss and a depth smoothness loss~\cite{godard2019digging},
\begin{equation}
L_{total}=L_{self}+ \lambda L_{smooth},
\end{equation}
where $\lambda$ is a weighting parameter and is set to $1.0\times10^{-3}$.  After the network is trained, we do inference for each coming image to predict its depth map and the uncertainty map, and feed them into our SLAM system.

\subsection{Uncertainty-Guided Visual-Inertial SLAM}
Our system is built upon ORB-SLAM3~\cite{campos2020orb}, but is compatible with any existing feature-based SLAM that supports RGB-D inertial input. Instead of relying a depth sensor to get the depth image, we use the CNN predicted depth as the depth input to form a pseudo RGB-D inertial SLAM. We further augment SLAM with the uncertainty prediction to adaptively weigh the contribution of feature points in the back-end optimization of bundle adjustment~\cite{triggs1999bundle}. 

Given map points $X^j\in\mathbb{R}^3$ in the world coordinates and the 2D observations $x_i^j\in\mathbb{R}^2$ in frame $i$, their re-projection errors can be computed as,
\begin{equation}
e_{ij}=x_i^j-\pi (T_i \bar{X}^j ),
\end{equation}
where $\pi$ is the projection function, $T_i\in\mathbb{R}^{3\times4}$ is the transformation from the world coordinate system to frame $i$, and $\bar{X}^j\in\mathbb{R}^4$ corresponds to the homogeneous coordinates of $X^j$. Bundle adjustment aims to minimize the reprojection errors of map points across a bundle of frames,
\begin{equation}
C^l=\sum_{i,j} \rho_h (e_{ij}^T \Omega_{ij}^{-1} e_{ij} ),
\end{equation}
where $\rho_h$ is the robust Huber cost function and $\Omega_{ij}$ is the covariance matrix measuring the uncertainty of each detected feature point. In a conventional geometric SLAM system, \eg, ORB-SLAM, the covariance is pre-computed according to the scale at which the feature point is detected; thus all feature points detected from the same scale share the same covariance. In this paper, we instead propose to use the CNN learned uncertainties to adaptively weigh the re-projection errors. Specifically, we use a simple method to convert the uncertainty value to a covariance metrix, \ie,
\begin{equation}
\Omega_{ij}=1/(I-U_{ij} ),
\end{equation}
where $U_{ij} = u_{ij} I_{2\times2}$.
Therefore, the uncertainty-guided reprojection errors become
\begin{equation}
C_{uncer}^l=\sum_{i,j} \rho_h (e_{ij}^T (I-U_{ij})e_{ij})).
\label{eq:uncer}
\end{equation}
Since our SLAM system takes CNN predicted depth as input and CNN depth is not always accurate, this uncertainty guidance help to downweigh the contributions of those map points that potentially have high depth errors.

Following ORB-SLAM3~\cite{mur2017orb}, we use the depth and left features to generate feature points on a virtual right view. We then compute the reprojection errors for the virtual right view so that we have the overall reprojection cost as follows,
\begin{equation}
    C_{reproj} = C_{uncer}^l + \alpha C_{uncer}^r,
\end{equation}
where $C_{uncer}^r$ is the reprojection cost for the virtual right view and $\alpha \leq 1$.

\subsection{Fast Plane Fitting}
The world is mostly made of planes, especially for indoor environments. At the application side, AR often requires generating virtual objects on top of real-world planar surfaces. So, in our system, we also make use of 3D planar information for back-end regularization. However, solely utilizing 3D map points suffers a lot from under-constrained situations, as many walls are texture-less. Instead, we draw upon both the dense CNN depth cue and sparse 3D map points.

It would be slow if we na\"ively fit planes using the dense depth map and all map points. We thus propose a novel fast plane fitting method, leveraging the prior information of gravity direction and uncertainty measurements. In particular, we seek to find two types of dominant planes: i) horizontal planes that are perpendicular to the gravity direction; ii) vertical planes that are parallel to the gravity direction.

In the general case where the gravity direction is unknown, we would have to sample three points to fit a plane. In contrast, given the gravity direction, the degree-of-freedom (DoF) of horizontal planes becomes one, so we only need to sample one point to determine a horizontal plane. Similarly, the DoF of vertical planes is two and we only need two points to determine a vertical plane. Note that, reduction in DoF of a parametric model can significantly speed up the RANSAC algorithm~\cite{fischler1981random}. Theoretically, the number of iterations for RANSAC equals 
\begin{equation}
k = \frac{\log(1-p)}{\log(1-w^n)},
\end{equation}
where $p$ is the desired success probability, $w$ is the inlier ratio, and $n$ is the number of points to fit the model. Assume $p=0.99$, $w=0.2$. If $n=3$, then $k=249$; if $n=2$, then $k=49$; if $n=1$, then $k=9$. From this example, we can see the benefits of applying the one-point RANSAC (for horizontal planes) and two-point RANSAC (for vertical planes) in terms of computational complexity, as compared to the vanilla three-point RANSAC.

We further assume that indoors scenes consist mainly of planes that are orthogonal to one another in a “Manhattan world” manner~\cite{coughlan1999manhattan,coughlan2003manhattan}. This assumption allows us to further accelerate the detection of wall planes. Once we have detected one main wall plane using two-point RANSAC method, we assume there are additional planes parallel or orthogonal to the main wall plane. This enables us to remove one DoF from the problem and test subsequent plane hypotheses with only one point per sample. 

When detecting planes from SLAM map points, we observe that false positive plane inliers such as points belonging to other surfaces can introduce errors in plane normal direction estimation. We also find that clusters of colinear points can often lead to false positive detection. We mitigate these challenges by augmenting map points with normal direction computed from CNN depth, which is computed as follows:
\begin{equation}
\begin{aligned}
n_{x,y} = & R\cdot(K^{-1}([x,y-\delta,1]^T - [x,y,1]^T)) \\
&\times(K^{-1}([x-\delta,y,1]^T - [x,y,1]^T)),
\end{aligned}
\end{equation}
where $n_{x,y}$ is the normal direction at image coordinate $(x,y)$, $K$ is the camera intrinsic matrix, $\delta$ stands for a small pixel shift where the corresponding surface is assumed to be piece-wise planar and $R$ represents the rotation matrix from the camera to the world coordinate system. Map point normals are maintained and updated with each CNN depth frame. Map point normals can help significantly reduce false positive inlier points, resulting in more robust plane detection especially for vertical planes.

Moreover, we observe that CNN depth often has a global scale shift. To overcome this issue, we additionally employ a simple yet effective method to refine CNN depth by using SLAM map points. As map points are continually optimized in the back-end and in general more accurate, it is therefore able to reproject them onto images to compute an average scale difference with CNN depth:
\begin{equation}
s_{correction} = \frac{1}{N}\sum_{i=1}^{N} \frac{D_i^{SLAM}}{D_i^{CNN}},
\end{equation}
where $N$ is the number of visible map points, $D_i^{SLAM}$ is the depth from the $i^{th}$ map point, and $D_i^{CNN}$ is the corresponding CNN depth.
Then, the computed factor is applied to CNN depth for scale correction.

We use the refined dense CNN depth map to assist recovering planes. Instead of equally sampling the CNN depth map, we utilize CNN uncertainties to guide the sampling process. In regions of higher uncertainties, the depth points are sampled with lower probability, and vice versa. The inlier count of CNN depth points is downweighed by $0.5$ as they are less accurate than SLAM map points. 

\begin{figure}[!t]
\centering
\begin{tikzpicture}
    \node (C0) at (0, 0) [circle, draw=black!100, fill=blue!20, text=black!100] {$T_0$};
    \node (C1) at (2, 0) [circle, draw=black!100, fill=blue!20, text=black!100] {$T_1$};
    \node (C3) at (4, 0) [circle, draw=black!100, fill=blue!20, text=black!100] {$T_3$};
    \node (Ct) at (7, 0) [circle, draw=black!100, fill=blue!20, text=black!100] {$T_t$};
    \node (Xvi) at (1, 2) [circle, draw=black!100, fill=yellow!20, text=black!100] {$X^{i}_{v}$};
    \node (Xj) at (6, 2) [circle, draw=black!100, fill=red!20, text=black!100] {$X^{j}$};
    \path coordinate (cent0) at (0.5, 1);
    \path coordinate (dot0) at (5, 0);
    \path coordinate (dot1) at (5.5, 0);
    \path coordinate (dot2) at (6, 0);
    \draw[-] (C0) to node[left] {$x_0$} (Xvi);
    \draw[-] (Xvi) to [in=120, out=60, loop] node[above] {0} (Xvi);
    \draw[-] (C1) to node[left] {$x_1$} (Xvi);
    \draw[-] (C3) to node[left] {$x_2$} (Xvi);
    \draw[-] (C3) to node[left] {$x_3$} (Xj);
    \draw[-] (Ct) to node[left] {$x_4$} (Xj);
    \fill[black] (0.4, 0.9) rectangle +(0.2, 0.2);
    \fill[black] (0.9, 3.1) rectangle +(0.2, 0.2);
    \fill[black] (1.4, 0.9) rectangle +(0.2, 0.2);
    \fill[black] (2.4, 0.9) rectangle +(0.2, 0.2);
    \fill[black] (4.9, 0.9) rectangle +(0.2, 0.2);
    \fill[black] (6.4, 0.9) rectangle +(0.2, 0.2);
    \fill[black] (dot0) circle (1.5pt);
    \fill[black] (dot1) circle (1.5pt);
    \fill[black] (dot2) circle (1.5pt);
\end{tikzpicture}
\caption{Factor graph for bundle adjustment with planar constraints. Note that our planar constraints are implemented as a unary term in the factor graph (\ie, a self-connecting edge for the node $X_v^i$).}
\label{fig:factor_graph}
\end{figure}
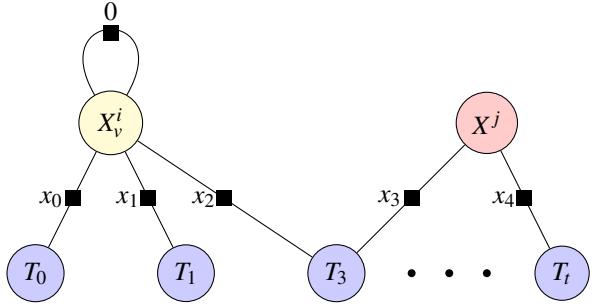

\subsection{Planar Constraints}

Given steadily detected planes, we build an extra energy term on top of the re-projection error term for back-end optimization. 
If a point has a high probability lying on a plane, the distance between them should be as small as possible. Therefore, for each detected plane $v \in \mathbb{R}^4$, we first find its inlying map points $X_v^j$ by thresholding the distance from map points to planes, which is controlled by $\theta$. Then the plane constraint term in the back-end optimization can be expressed as:
\begin{equation}
C_{plane} = \sum_{j,v} \rho_h(v^T\bar{X}_v^j),
\end{equation}
where $\bar{X}_v^j$ corresponds to the homogeneous coordinates of $X_v^j$. Since the plane types used in our work, such as walls and the ground, are well defined, it is not necessary to optimize plane parameters along with other variances in the back-end. Instead, we adopt an iterative manner, where planes are recomputed asynchronously in every $N_d$ frames. We set $N_d = 30$ in all of our experiments. Then the newly updated inlying planar points and plane parameters are used in the following graph optimization. In this manner, we can effectively express the planar constraint as a serial of unary edges in the factor graph optimization~\cite{kummerle2011g}, as shown in Figure~\ref{fig:factor_graph}. The observations on these edges are set to $0$, as theoretically the residual between planes and inlying points should be $0$. Our overall cost for bundle adjustment is the summation of $C_{reproj}$ and $C_{plane}$, \ie,
\begin{equation}
    C = C_{reproj} + \beta C_{plane},
\end{equation}
where $\beta \leq 1$, and is set to $0.1$ in our experiments. The planar constraint can be applied in both local and full bundle adjustment processes as a guidance for map points optimization.

\section{EXPERIMENTS}

We evaluate our system on the EuRoC MAV dataset~\cite{Burri25012016} which provides stereo images, IMU data, and groundtruth camera 6-DoF poses. We choose the VICON Room sequences of EuRoC to perform the evaluation because these sequences contain large planes, such as the ground plane and walls. We exclude Sequence V203 due to many missing frames in one of the cameras~\cite{yang2020d3vo}. In the following subsections, we first discuss the implementation details of our system, and then present both quantitative and qualitative results of ORB-SLAM3~\cite{campos2020orb} and our method on EuRoC.

\subsection{Implementation Details}

\begin{figure}[!t]
\centering
\includegraphics[width=0.48\textwidth]{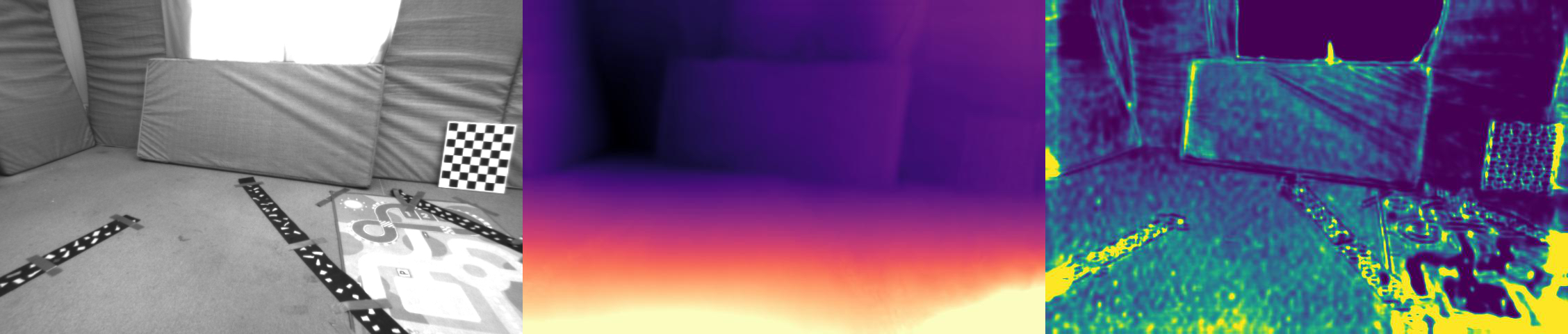} \\
\includegraphics[width=0.48\textwidth]{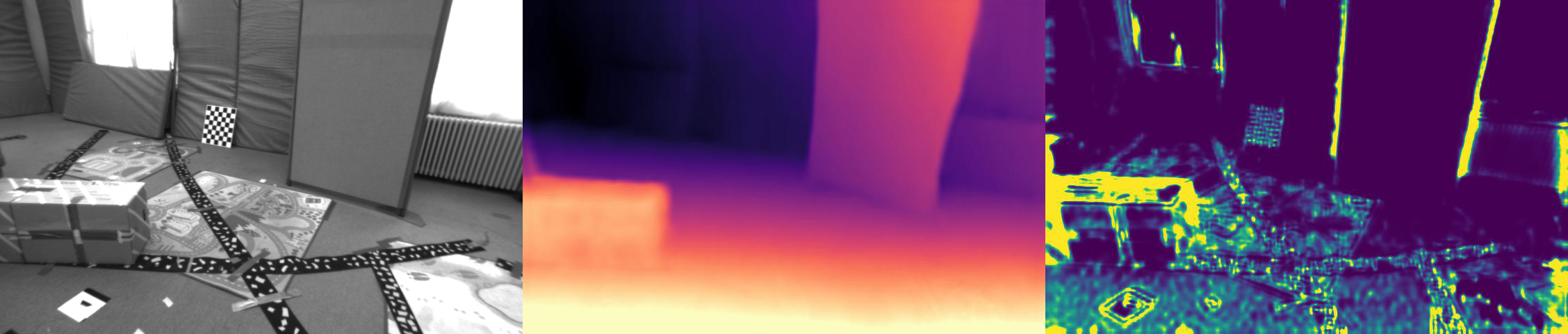} \\
\includegraphics[width=0.48\textwidth]{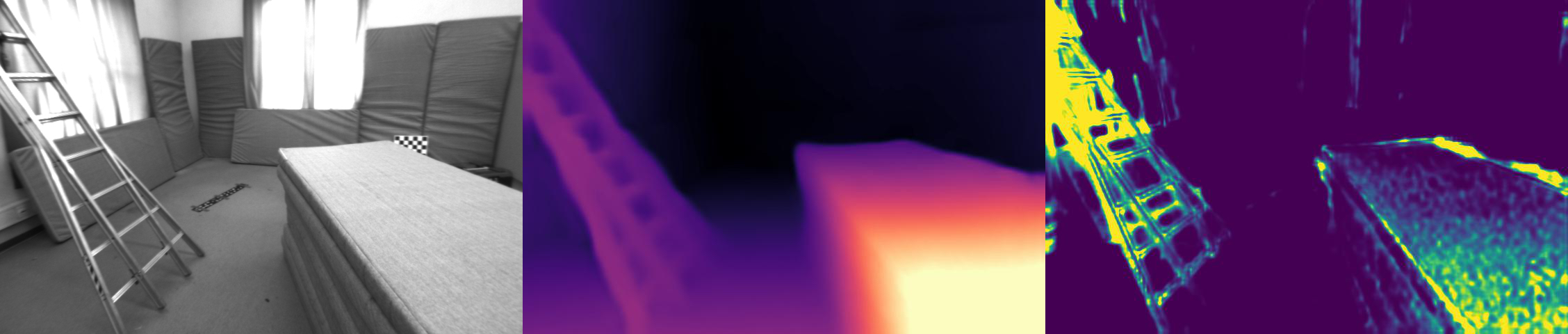}
\vspace{-0.5cm}
\caption{Depth and uncertainty prediction. From the left to the right are respectively the original input, the CNN depth and the CNN uncertainty.}
\label{fig:vis_compare}
\end{figure}

We follow~\cite{yang2020d3vo} to train the network to predict the depth and uncertainty using monocular and stereo images in a self-supervised manner. Due to the limited size of this dataset, we use five VICON Room sequences for training to achieve the best model. A supervised method can also be used to predict the depth map. The depth and uncertainty maps of some sample images are shown in Figure~\ref{fig:cnn-depth-uncer}. We replace the covariance matrix in the back-end bundle adjustment with the CNN uncertainty prediction as in Eq.~\eqref{eq:uncer} and set $\alpha$ to 1. To better gauge the drift in camera poses, we disable loop closure for all competing methods.

In plane detection, map point surface normal estimated from CNN depth are used to help filter out outliers. We maintain a hash map from map point ID to their surface normals as well as their last observed distance. Surface normals of map points in view are updated by new CNN depth frames. The new normal vector is a weighted average based on inverse of observation distance,
\begin{equation}
\overrightarrow{n} = \frac{\overrightarrow{n}_{new}\cdot{d} + \overrightarrow{n}\cdot{d_{new}}}{d_{new}+d},
\end{equation}
where $\overrightarrow{n}$ is the estimated map point surface normal, $d$ is the observation distance of the map point and subscript $new$ indicates new values estimated from the current frame.

We implement multi-plane detection by iteratively detecting the max inlier count plane, recording its inlier points and detecting the next plane from the remaining points. We found the inlier distance threshold $\theta = 0.05 m$ and ratio $w = 0.1$ work well for VICON sequences of the EuRoC dataset.


\subsection{Experimental Results}

\begin{table}[!t]
\caption{Performance comparison on EuRoC (RMSE ATE in meters).}
\vspace{-0.4cm}
\begin{center}
\scalebox{0.89}{
\begin{tabular}{|p{1.8cm}|c|c|c|c|c|c|}
\hline
& V101 & V102 & V103 & V201 & V202 & Mean \\
\hline
\hline
\begin{tabular}{@{}c@{}}ORB-SLAM3 \\ Mono-Inertial\end{tabular}       & 0.0367       & 0.0605 & 0.0767 & 0.0480 & 0.0620 & 0.0568 \\
\hline
\begin{tabular}{@{}c@{}}ORB-SLAM3 \\ Stereo-Inertial\end{tabular}     & {\bf 0.0293} & 0.0596 & 0.0729 & 0.0699 & 0.0637 & 0.0591\\ 
\hline
\hline
\begin{tabular}{@{}c@{}}Our Mono-CNN \\ Inertial\end{tabular}         & 0.0352       & 0.0151 & 0.0381 & 0.0320 & 0.0404 & 0.0322\\
\hline
\begin{tabular}{@{}c@{}}Our Mono-CNN \\ Inertial + P\end{tabular}     & 0.0346  &  0.0143 & 0.0352 & 0.0315 & {\bf 0.0393} & 0.0310 \\
\hline
\begin{tabular}{@{}c@{}}Our Mono-CNN \\ Inertial + U\end{tabular}     & 0.0355   &  0.0127 & 0.0315 &  0.0244 & 0.0583 & 0.0325\\
\hline
\begin{tabular}{@{}c@{}}Our Mono-CNN \\ Inertial + U + P\end{tabular} &  0.0352      & {\bf 0.0126} & {\bf 0.0283} & {\bf 0.0234} & 0.0468 & {\bf 0.0293}\\ 
\hline
\end{tabular}
}
\end{center}
\label{table:noloop}
\end{table}

Following~\cite{campos2020orb}, we use the RMSE absolute translation error (ATE) as the quantitative evaluation metric. We compare our method with both the mono-inertial mode and the stereo-inertial mode of ORB-SLAM3, and show the performance comparison in Table~\ref{table:noloop}, where all numbers are computed by taking the median of 5 runs. We note that the mono-inertial mode and the stereo-inertial mode of ORB-SLAM3 work on par with each other. To ablate the contribution of each new component, we add one new module each time to the baseline system (\ie, ORB-SLAM3 Mono-Inertial), leading to 
four versions of our system, denoted as Our Mono-CNN Inertial, Our Mono-CNN Inertial + P(lane), Our Mono-CNN Inertial + U(ncertainty), and Our Mono-CNN Inertial + U(ncertainty) + P(lane), respectively. By comparing Our Mono-CNN Inertial and ORB-SLAM3 Mono-Inertial in Table~\ref{table:noloop}, it is evident that CNN depth consistently improves the accuracy of camera poses in all five sequences, and in some cases by a large margin (\eg, for V102). This verifies the contribution of CNN depth in bootstraping the performance of feature-based SLAM. Adding uncertainty prediction into the system (\ie, Our Mono-CNN Inertial + U) further improves the performance in 3 out of 5 sequences. Comparing Our Mono-CNN Inertial + P and Our Mono-CNN Inertial + U + P, we can also observe that the mean RMSE of ATE improves by using CNN uncertainty. Our final system (\ie, Our Mono-CNN Inertial + U + P) consistently outperforms Our Mono-CNN Inertial + U and achieves the best performance, combining the benefits of CNN depth, uncertainty, and structural constraints.

\begin{figure}[!htb]
\centering
\vspace{-0.2cm}
\includegraphics[width=0.45\textwidth]{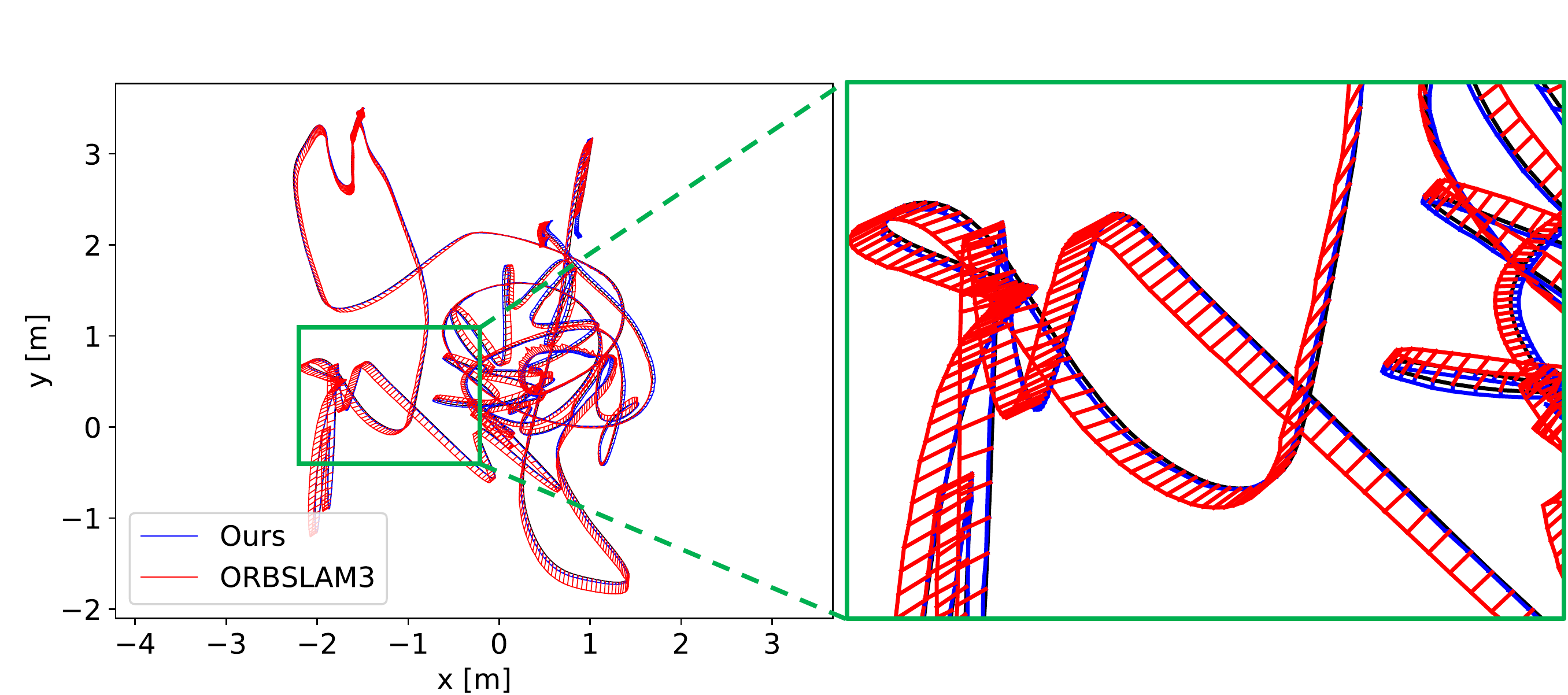} \\
\includegraphics[width=0.45\textwidth]{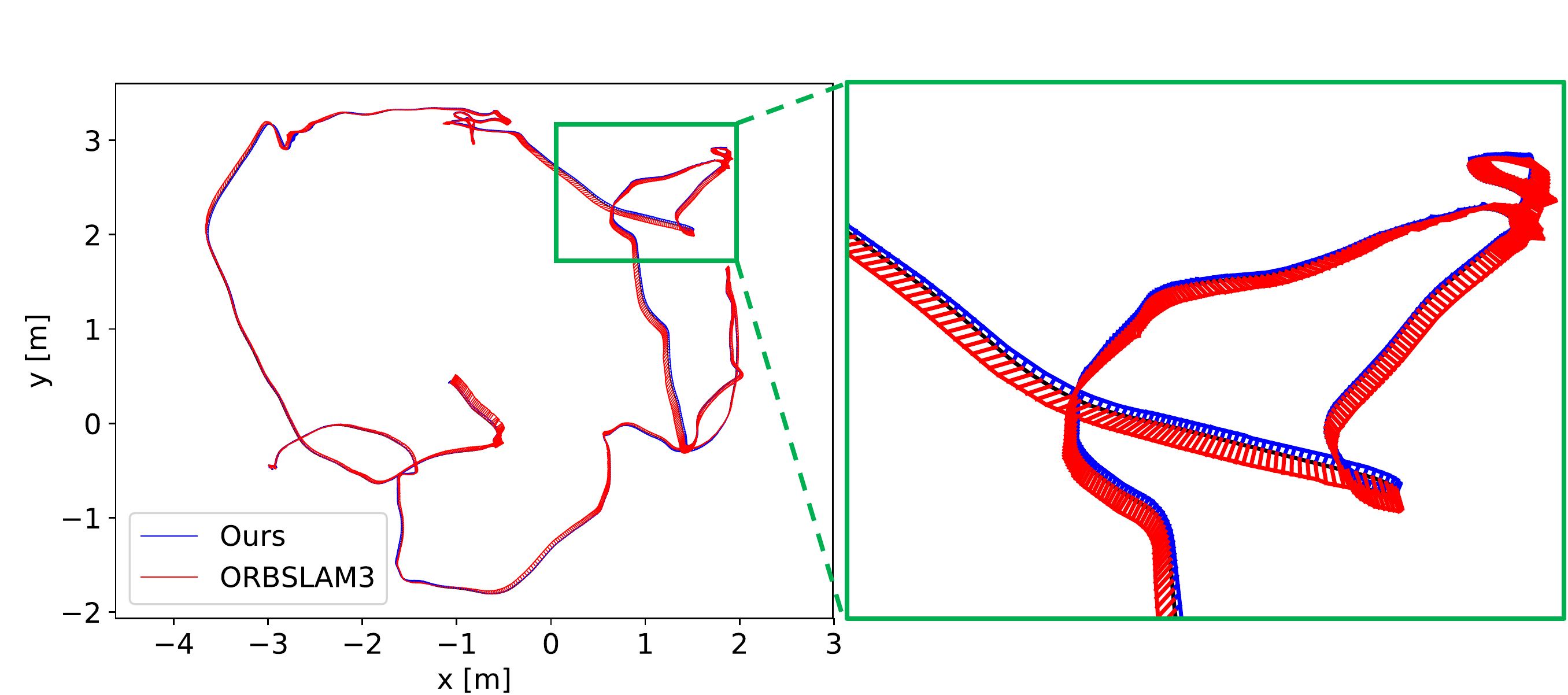} \\
\includegraphics[width=0.45\textwidth]{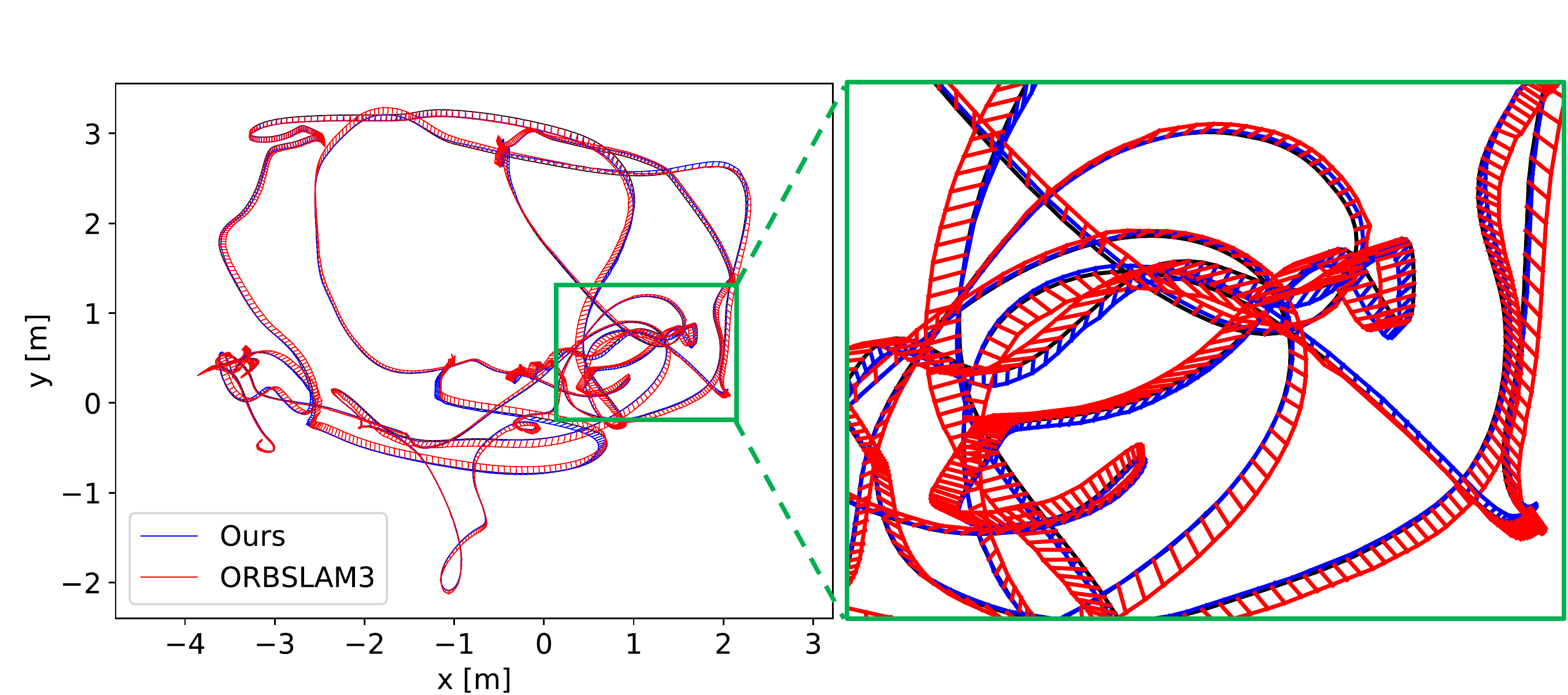}
\vspace{-0.2cm}
\caption{Camera trajectory visual comparison on EuRoC VICON sequences. We plot the difference between the ORB-SLAM3 trajectory and the groundtruth in red and the difference between our trajectory and the groundtruth in blue. We can clearly see that our trajectories align better with the groundtruth than ORB-SLAM3. Best viewed on screen with zoom-in.}
\label{fig:vis_trajectory}
\end{figure}

To investigate the stability of our system, we show the boxplot of the RMSE values of ATE over 5 runs for all the 5 sequences in Figure~\ref{fig:vis_trajectory}. We can observe that, on V101 and V102, our system not only achieves lower the RMSE ATE than ORB-SLAM3, but is also more stable with smaller variance. On V103 and V201, our system is on par with ORB-SLAM3 in terms of performance stability. However, our system demonstrates larger variance on V202. We conjecture that this is due to much more violent motions in V202 which cause difficulties in self-supervised depth and uncertainty prediction~\cite{yang2020d3vo}.

\begin{figure}[!t]
\centering
\begin{tabular}{cc}
\includegraphics[width=0.25\textwidth]{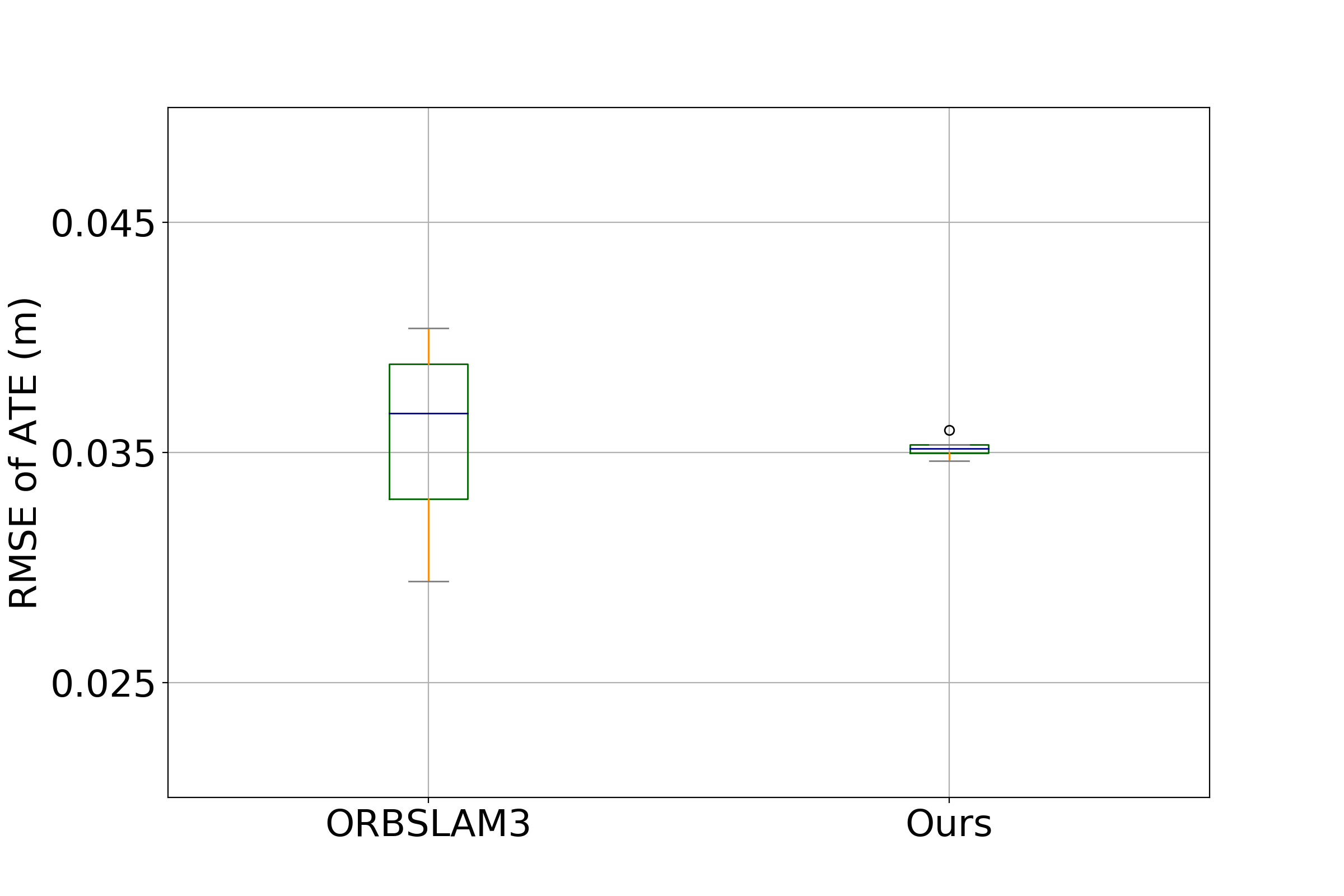} 
&
\hspace{-0.8cm}
\includegraphics[width=0.25\textwidth]{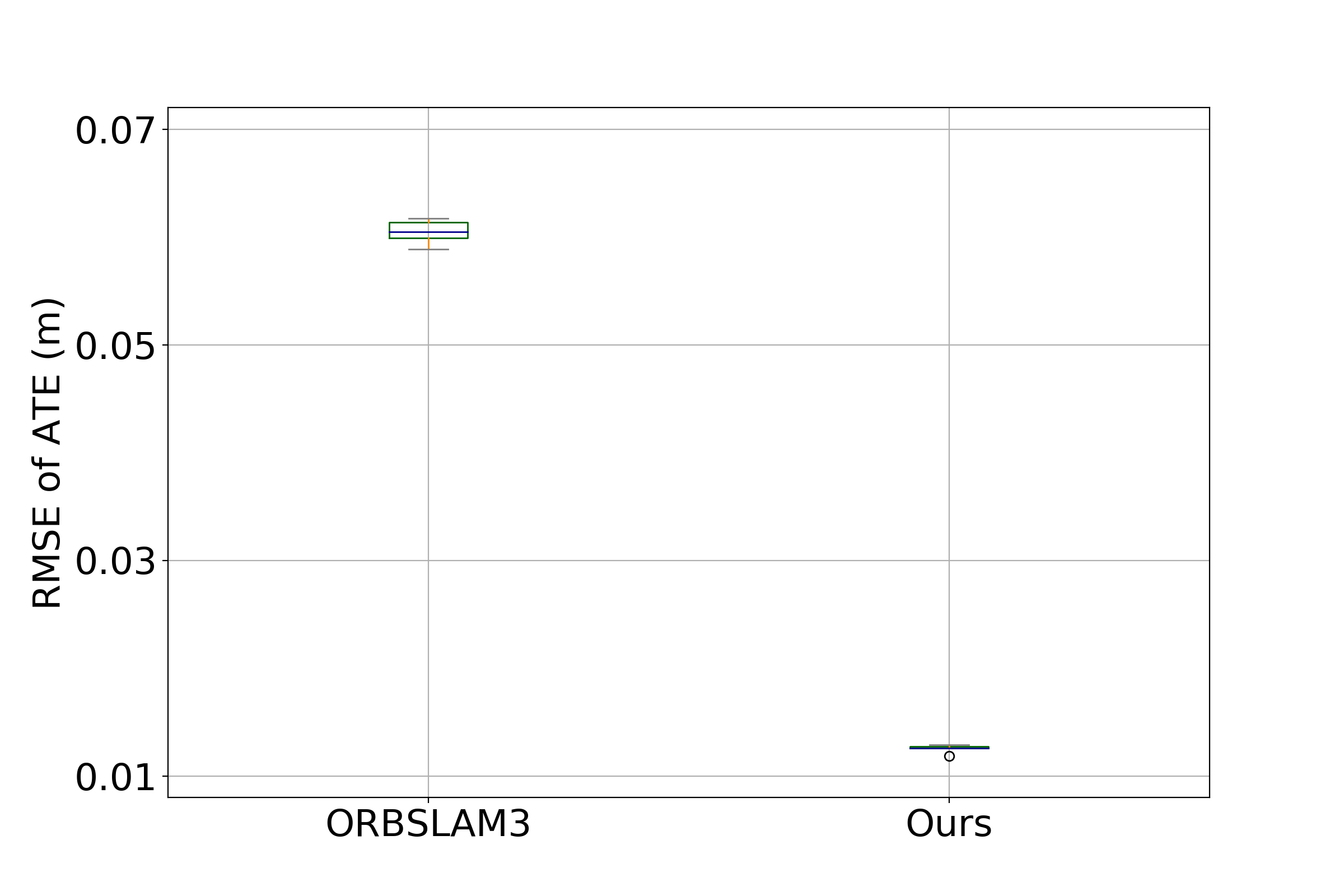} \vspace{-0.3cm}\\
{\scriptsize V101} & \hspace{-0.8cm} {\scriptsize V102} \vspace{-0.1cm}\\
\includegraphics[width=0.25\textwidth]{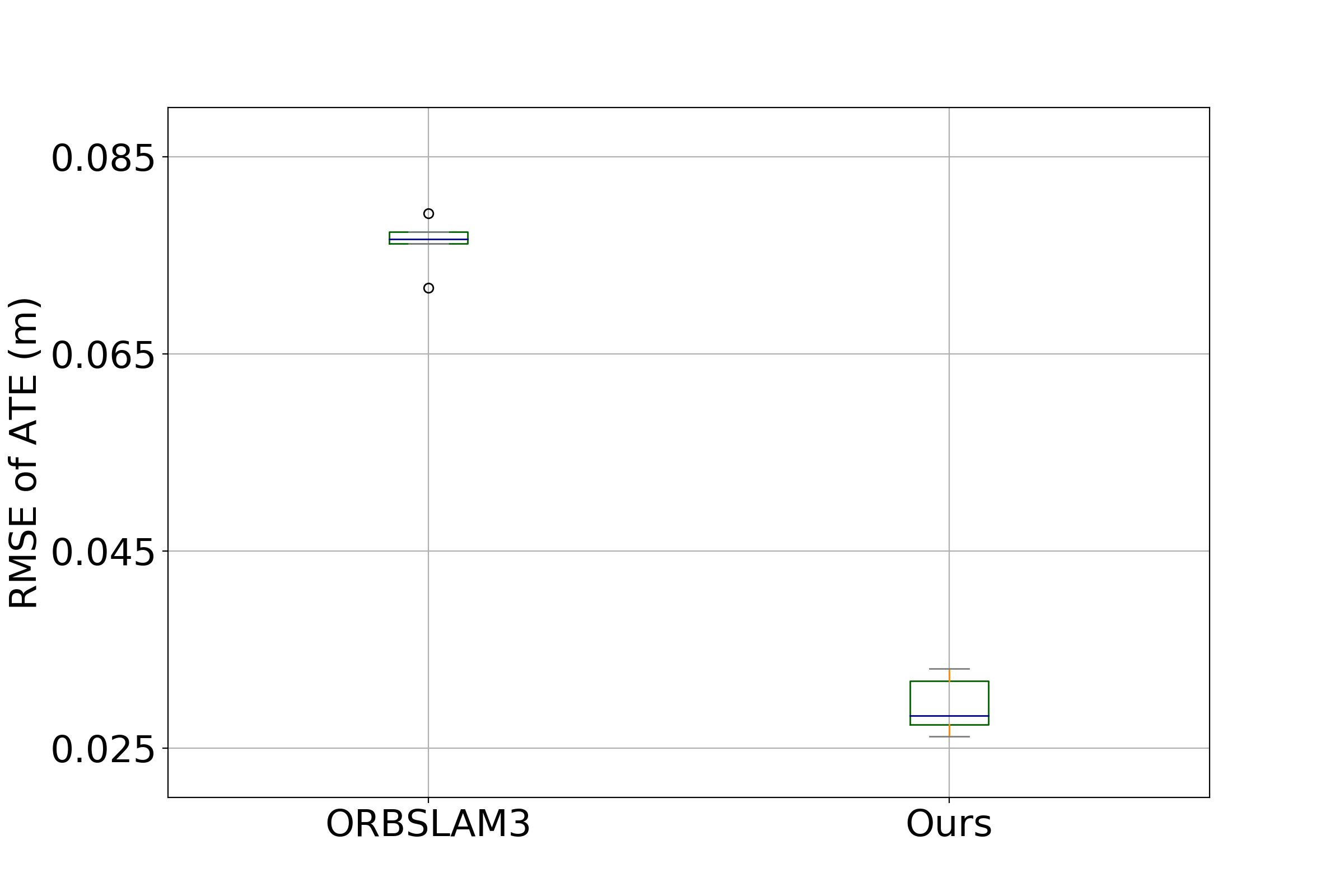} 
&
\hspace{-0.8cm}
\includegraphics[width=0.25\textwidth]{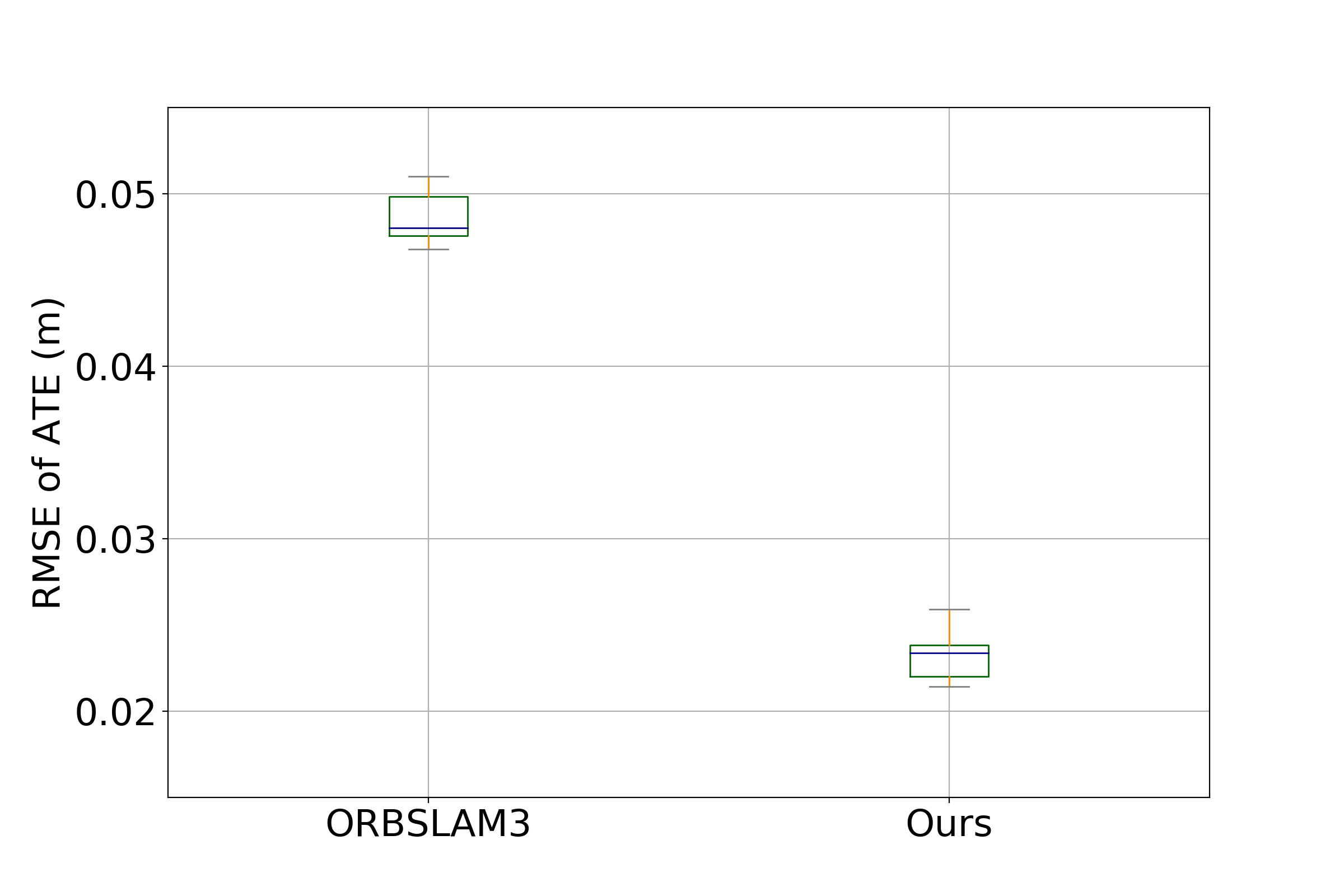} \vspace{-0.3cm}\\
{\scriptsize V103} & \hspace{-0.8cm} {\scriptsize V201} \vspace{-0.1cm} \\
& \hspace{-5.0cm}\includegraphics[width=0.25\textwidth]{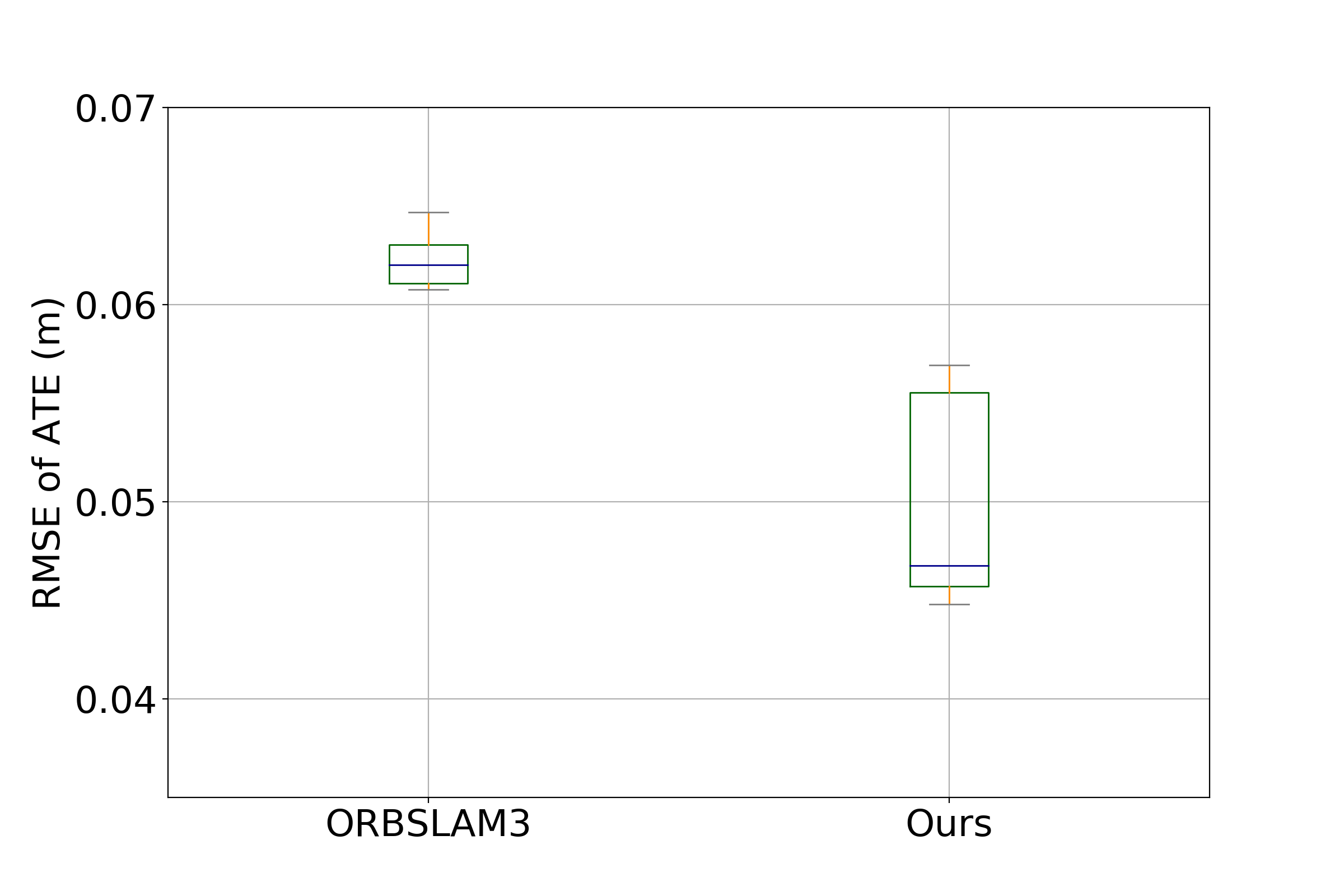} \vspace{-0.3cm}\\
& \hspace{-5.0cm} {\scriptsize V202} \\
\end{tabular}
\caption{Boxplot of the RMSE values of ATE over 5 runs on VICON sequences. Best viewed on screen with zoom-in.}
\label{fig:boxplot}
\end{figure}

In Figure~\ref{fig:vis_trajectory}, we further give a visual comparison of camera trajectories computed by ORB-SLAM3 and our system. From the figure, it is clear that our system recovers more accurate camera poses.

\begin{figure}[!t]
\centering
\begin{tabular}{cc}
\includegraphics[width=0.2\textwidth]{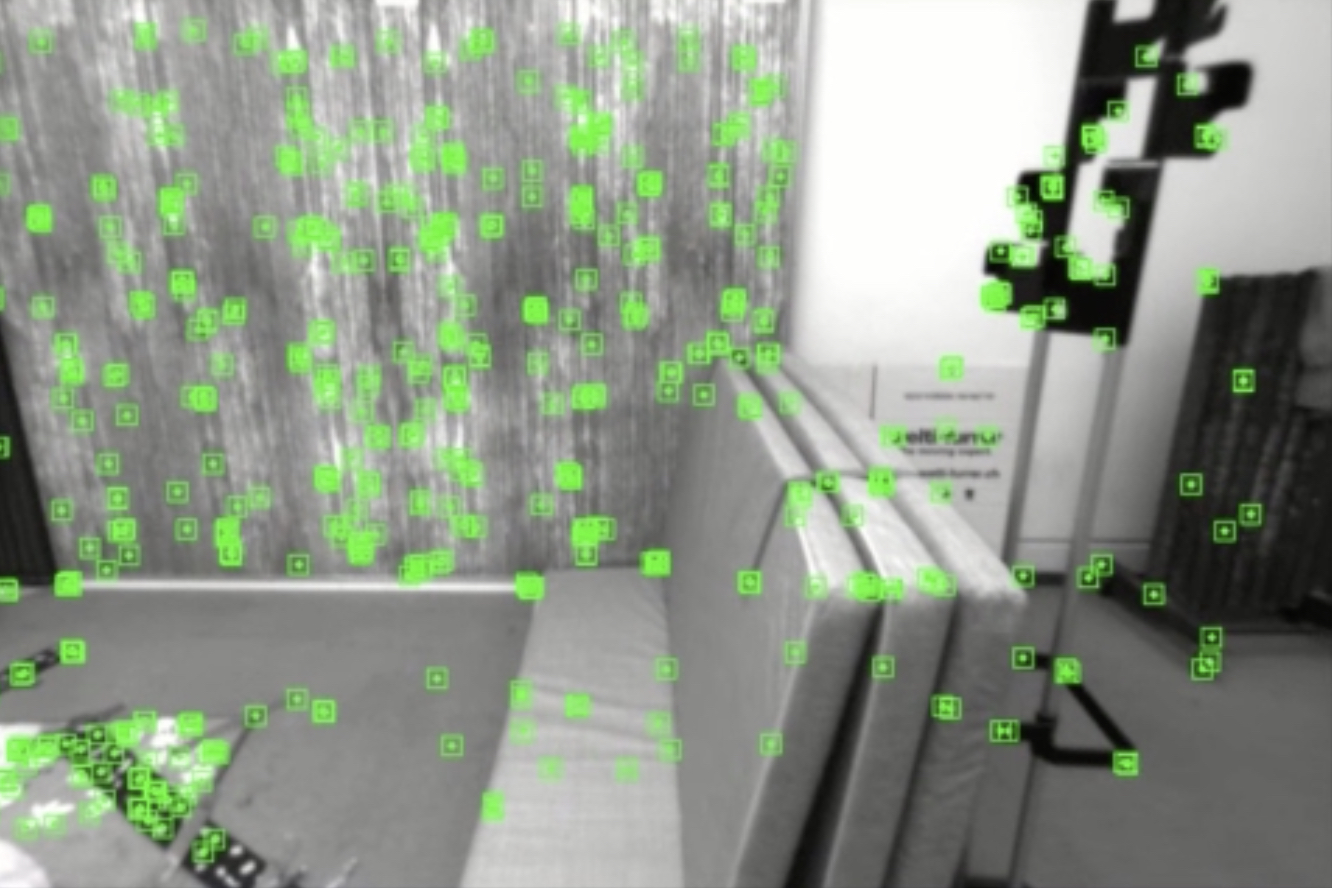} 
\includegraphics[width=0.2\textwidth]{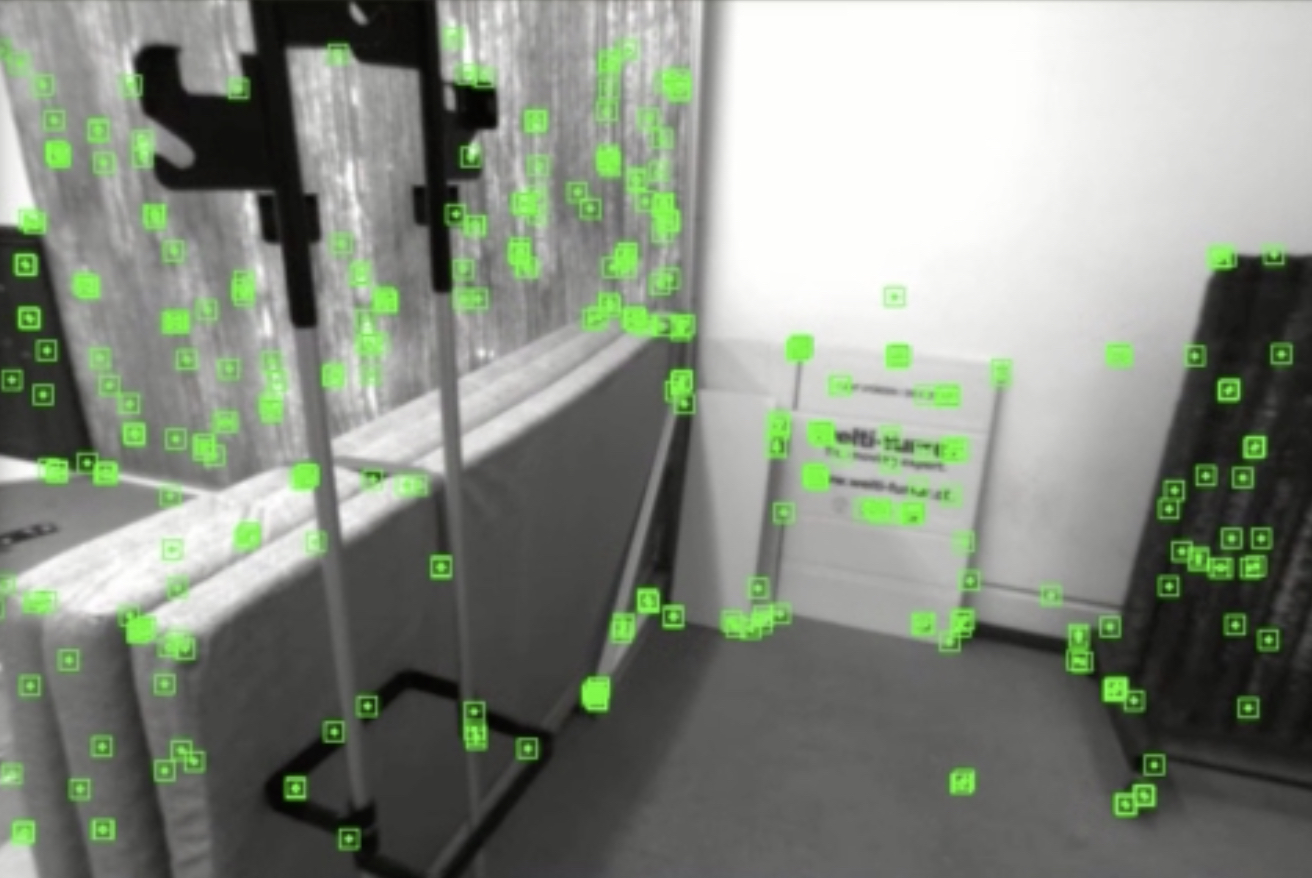} \\
\includegraphics[width=0.38\textwidth]{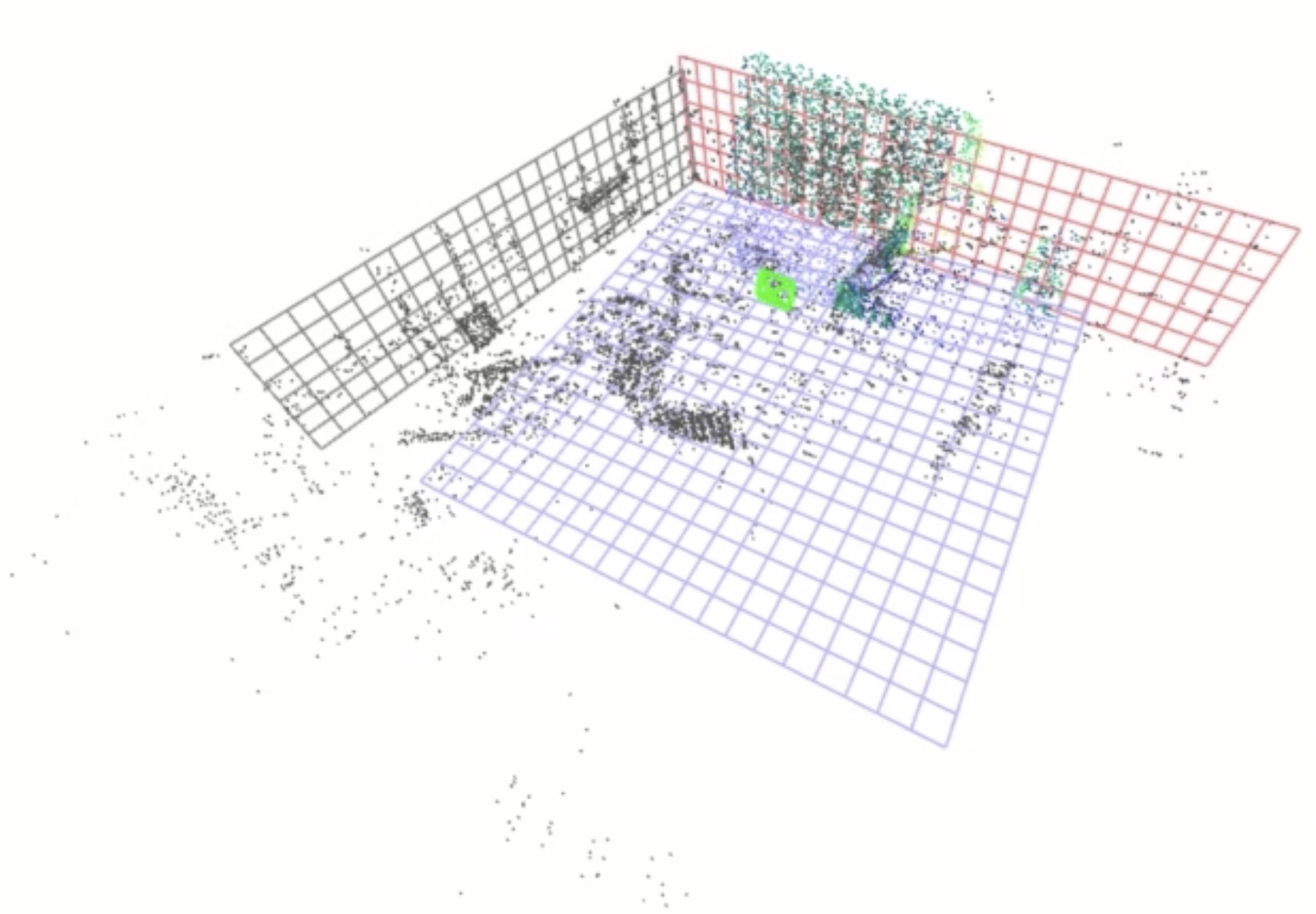} 
\end{tabular}
\vspace{-0.7cm}
\caption{Plane visualization. The horizontal ground plane is plotted in blue, the vertical planes are in red, and the invisible planes (in the current time step) are in black.}
\label{fig:plane_vis}
\end{figure}

Our system can detect and fit multiple dominant planes that are present in the scene. We visualize the plane detection in a sample scene in Figure~\ref{fig:plane_vis}. Our system is able to detect and find 3D horizontal planes and vertical planes reliably. We also encourage the reader to watch the supplementary demo video for our plane detection.

\section{CONCLUSIONS}
The proposed system maintains the minimal sensor setup of using a monocular camera and an inertial sensor (\eg, IMU) for SLAM, yet achieving the benefits of using the RGB-D inertial input via a CNN to predict depth and uncertainty. In this sense, the proposed system combines the best of both worlds (\ie, geometry and deep learning). In addition, we present a fast and effective plane detection method along with our SLAM system. Our system is fast (real-time) and accurate in reconstructing map points, camera poses, and major planes. 

{\small
\bibliographystyle{ieee}
\bibliography{slam.bib}

\begin{thebibliography}{10}\itemsep=-1pt

\bibitem{arndt2020points}
C.~Arndt, R.~Sabzevari, and J.~Civera.
\newblock From points to planes-adding planar constraints to monocular slam
  factor graphs.
\newblock In {\em IROS}, 2020.

\bibitem{Burri25012016}
M.~Burri, J.~Nikolic, P.~Gohl, T.~Schneider, J.~Rehder, S.~Omari, M.~W.
  Achtelik, and R.~Siegwart.
\newblock The euroc micro aerial vehicle datasets.
\newblock {\em The International Journal of Robotics Research}, 2016.

\bibitem{cadena2016past}
C.~Cadena, L.~Carlone, H.~Carrillo, Y.~Latif, D.~Scaramuzza, J.~Neira, I.~Reid,
  and J.~J. Leonard.
\newblock Past, present, and future of simultaneous localization and mapping:
  Toward the robust-perception age.
\newblock {\em IEEE Transactions on robotics}, 32(6):1309--1332, 2016.

\bibitem{campos2020orb}
C.~Campos, R.~Elvira, J.~J.~G. Rodr{\'\i}guez, J.~M. Montiel, and J.~D.
  Tard{\'o}s.
\newblock {ORB-SLAM3}: An accurate open-source library for visual,
  visual-inertial and multi-map slam.
\newblock {\em arXiv preprint arXiv:2007.11898}, 2020.

\bibitem{coughlan1999manhattan}
J.~Coughlan and A.~Yuille.
\newblock Manhattan world: Compass direction from a single image by bayesian
  inference.
\newblock In {\em IEEE International Conference on Computer Vision}, 1999.

\bibitem{coughlan2003manhattan}
J.~Coughlan and A.~Yuille.
\newblock Manhattan world.
\newblock {\em Neural Computation}, 15:1063--1088, 2003.

\bibitem{engel2017direct}
J.~Engel, V.~Koltun, and D.~Cremers.
\newblock Direct sparse odometry.
\newblock {\em TPAMI}, 40(3):611--625, 2017.

\bibitem{engel2014lsd}
J.~Engel, T.~Sch{\"o}ps, and D.~Cremers.
\newblock {LSD-SLAM}: Large-scale direct monocular slam.
\newblock In {\em ECCV}, 2014.

\bibitem{fischler1981random}
M.~A. Fischler and R.~C. Bolles.
\newblock Random sample consensus: a paradigm for model fitting with
  applications to image analysis and automated cartography.
\newblock {\em Communications of the ACM}, 24(6):381--395, 1981.

\bibitem{fu2018deep}
H.~Fu, M.~Gong, C.~Wang, K.~Batmanghelich, and D.~Tao.
\newblock Deep ordinal regression network for monocular depth estimation.
\newblock In {\em CVPR}, 2018.

\bibitem{godard2019digging}
C.~Godard, O.~Mac~Aodha, M.~Firman, and G.~J. Brostow.
\newblock Digging into self-supervised monocular depth estimation.
\newblock In {\em ICCV}, 2019.

\bibitem{hosseinzadeh2018structure}
M.~Hosseinzadeh, Y.~Latif, T.~Pham, N.~Suenderhauf, and I.~Reid.
\newblock Structure aware slam using quadrics and planes.
\newblock In {\em ACCV}, 2018.

\bibitem{hsiao2018dense}
M.~Hsiao, E.~Westman, and M.~Kaess.
\newblock Dense planar-inertial slam with structural constraints.
\newblock In {\em ICRA}, 2018.

\bibitem{ji2021monoindoor}
P.~Ji, R.~Li, B.~Bhanu, and Y.~Xu.
\newblock Monoindoor: Towards good practice of self-supervised monocular depth
  estimation for indoor environments.
\newblock In {\em ICCV}, pages 12787--12796, 2021.

\bibitem{ji2022georefine}
P.~Ji, Q.~Yan, Y.~Ma, and Y.~Xu.
\newblock Georefine: Self-supervised online depth refinement for accurate dense
  mapping.
\newblock {\em arXiv preprint arXiv:2205.01656}, 2022.

\bibitem{kaess2015simultaneous}
M.~Kaess.
\newblock Simultaneous localization and mapping with infinite planes.
\newblock In {\em ICRA}, 2015.

\bibitem{kendall2017uncertainties}
A.~Kendall and Y.~Gal.
\newblock What uncertainties do we need in bayesian deep learning for computer
  vision?
\newblock In {\em NuerIPS}, 2017.

\bibitem{kummerle2011g}
R.~K{\"u}mmerle, G.~Grisetti, H.~Strasdat, K.~Konolige, and W.~Burgard.
\newblock g 2 o: A general framework for graph optimization.
\newblock In {\em ICRA}, 2011.

\bibitem{lee2011mav}
G.~H. Lee, F.~Fraundorfer, and M.~Pollefeys.
\newblock Mav visual slam with plane constraint.
\newblock In {\em ICRA}, 2011.

\bibitem{liu2022planemvs}
J.~Liu, P.~Ji, N.~Bansal, C.~Cai, Q.~Yan, X.~Huang, and Y.~Xu.
\newblock Planemvs: 3d plane reconstruction from multi-view stereo.
\newblock {\em arXiv preprint arXiv:2203.12082}, 2022.

\bibitem{loo2019cnn}
S.~Y. Loo, A.~J. Amiri, S.~Mashohor, S.~H. Tang, and H.~Zhang.
\newblock {CNN-SVO}: Improving the mapping in semi-direct visual odometry using
  single-image depth prediction.
\newblock In {\em ICRA}, 2019.

\bibitem{mur2015orb}
R.~Mur-Artal, J.~M.~M. Montiel, and J.~D. Tardos.
\newblock {ORB-SLAM}: a versatile and accurate monocular slam system.
\newblock {\em IEEE transactions on robotics}, 31(5):1147--1163, 2015.

\bibitem{mur2017orb}
R.~Mur-Artal and J.~D. Tard{\'o}s.
\newblock {ORB-SLAM2}: An open-source slam system for monocular, stereo, and
  rgb-d cameras.
\newblock {\em IEEE Transactions on Robotics}, 33(5):1255--1262, 2017.

\bibitem{qin2018vins}
T.~Qin, P.~Li, and S.~Shen.
\newblock Vins-mono: A robust and versatile monocular visual-inertial state
  estimator.
\newblock {\em IEEE Transactions on Robotics}, 34(4):1004--1020, 2018.

\bibitem{song2013parallel}
S.~Song, M.~Chandraker, and C.~C. Guest.
\newblock Parallel, real-time monocular visual odometry.
\newblock In {\em ICRA}, 2013.

\bibitem{tateno2017cnn}
K.~Tateno, F.~Tombari, I.~Laina, and N.~Navab.
\newblock {CNN-SLAM}: Real-time dense monocular slam with learned depth
  prediction.
\newblock In {\em CVPR}, 2017.

\bibitem{tiwari2020pseudo}
L.~Tiwari, P.~Ji, Q.-H. Tran, B.~Zhuang, S.~Anand, and M.~Chandraker.
\newblock Pseudo rgb-d for self-improving monocular slam and depth prediction.
\newblock In {\em ECCV}, 2020.

\bibitem{triggs1999bundle}
B.~Triggs, P.~F. McLauchlan, R.~I. Hartley, and A.~W. Fitzgibbon.
\newblock Bundle adjustment—a modern synthesis.
\newblock In {\em International workshop on vision algorithms}, pages 298--372.
  Springer, 1999.

\bibitem{yang2020d3vo}
N.~Yang, L.~v. Stumberg, R.~Wang, and D.~Cremers.
\newblock {D3VO}: Deep depth, deep pose and deep uncertainty for monocular
  visual odometry.
\newblock In {\em CVPR}, 2020.

\bibitem{yang2018deep}
N.~Yang, R.~Wang, J.~Stuckler, and D.~Cremers.
\newblock Deep virtual stereo odometry: Leveraging deep depth prediction for
  monocular direct sparse odometry.
\newblock In {\em ECCV}, 2018.

\bibitem{yang2016pop}
S.~Yang, Y.~Song, M.~Kaess, and S.~Scherer.
\newblock Pop-up {SLAM}: Semantic monocular plane slam for low-texture
  environments.
\newblock In {\em IROS}, 2016.

\bibitem{zhou2020efficient}
L.~Zhou, D.~Koppel, H.~Ju, F.~Steinbruecker, and M.~Kaess.
\newblock An efficient planar bundle adjustment algorithm.
\newblock In {\em ISMAR}, 2020.

\bibitem{zou2020learning}
Y.~Zou, P.~Ji, Q.-H. Tran, J.-B. Huang, and M.~Chandraker.
\newblock Learning monocular visual odometry via self-supervised long-term
  modeling.
\newblock {\em arXiv preprint arXiv:2007.10983}, 2020.

\end{thebibliography}
}

\end{document}